\documentclass[letterpaper]{article} 
\usepackage[preprint]{aaai2027}  
\usepackage[hyphens]{url}  
\usepackage{graphicx} 
\usepackage{natbib}  
\usepackage{caption} 
\usepackage{algorithm}
\usepackage[noend]{algpseudocode}

\usepackage{newfloat}
\usepackage{listings}
\DeclareCaptionStyle{ruled}{labelfont=normalfont,labelsep=colon,strut=off} 
\floatstyle{ruled}
\newfloat{listing}{tb}{lst}{}
\floatname{listing}{Listing}

\usepackage{booktabs}
\usepackage{xcolor}   
\usepackage{float}
\usepackage{pgfplots}
\pgfplotsset{compat=1.18}
\usepackage{subcaption}
\usepackage{amsmath}
\usepackage{amssymb}
\usepackage{mathtools}
\usepackage{booktabs}
\usepackage{amsthm}
\usepackage{tikz}
\usepackage{centernot}
\usetikzlibrary{arrows.meta}
\usepackage{xspace}
\usepackage{enumitem}

\usetikzlibrary{spn}
\usepackage[dvipsnames,svgnames]{xcolor}
\definecolor{mygrey}{gray}{0.8}

\theoremstyle{plain}
\newtheorem{theorem}{Theorem}

\newtheorem{proposition}{Proposition}
\newtheorem{lemma}{Lemma}
\theoremstyle{definition}
\newtheorem{definition}{Definition}

\theoremstyle{remark}

\newtheoremstyle{case}{}{}{}{}{}{:}{ }{}
\theoremstyle{case}

\makeatletter
\newenvironment{proofsketch}[1][\proofsketchname]{\par
  \pushQED{\qed}%
  \normalfont \topsep6\p@\@plus6\p@\relax
  \trivlist
  \item[\hskip\labelsep \itshape #1\@addpunct{.}]\ignorespaces
}{%
  \popQED\endtrivlist\@endpefalse
}
\newcommand{\proofsketchname}{Proof sketch}
\makeatother

\newcommand{\br}[1]{{\{#1\}}}

\newcommand{\TP}{\Tilde{P}}
\newcommand{\tiltheta}{\Tilde{\theta}}

\newcommand{\KL}[2]{D_{\mathsf{KL}}(#1 \Vert #2)}
\newcommand{\Dkl}[2]{D_{\mathsf{KL}}(#1 \Vert #2)}
\newcommand{\Dtv}[2]{D_{\mathsf{TV}}(#1 \Vert #2)}
\newcommand{\Dh}[2]{D_{\mathsf{H}}(#1 \Vert #2)}

\newcommand{\abs}[1]{\left\lvert#1\right\rvert}

\newcommand{\rvars}[1]{\ensuremath{\mathbf{#1}}\xspace}
\newcommand{\X}{\rvars{X}}
\newcommand{\Z}{\rvars{Z}}
\newcommand{\E}{\mathbb{E}}
\newcommand{\Y}{\rvars{Y}}

\newcommand{\C}{\mathcal{C}}

\newcommand{\I}{\mathbb{I}}

\newcommand{\supp}{\mathsf{supp}}
\newcommand{\ch}[0]{\mathsf{ch}}

\newcommand{\tree}[0]{\mathsf{tree}}

\newcommand{\jstate}[1]{\ensuremath{\mathbf{#1}}\xspace}
\newcommand{\x}{\jstate{x}}
\newcommand{\z}{\jstate{z}}
\newcommand{\y}{\jstate{y}}

\newcommand{\midlinewidth}{1.0pt}

\title{Statistical Guarantees for Probabilistic Circuit Parameter Learning}
\author {
    John Leland\textsuperscript{\rm 1},
    YooJung Choi\textsuperscript{\rm 1}
}
\affiliations {
    \textsuperscript{\rm 1}School of Computing and Augmented Intelligence\\
    Arizona State University\\
    Tempe, Arizona, USA\\
    jslelan1@asu.edu, yj.choi@asu.edu
}

\begin{document}

\maketitle

\begin{abstract}
    Probabilistic circuits (PCs) are a powerful class of models that balance both expressivity and tractable inference. The expressive efficiency of PC classes has been studied extensively in the exact case and, although to a lesser degree, the approximate case. However, these results are mainly concerned with the theoretical ability of circuits to compactly represent functions. Statistical guarantees, such as sample complexity, for learning PCs has remained largely unexplored. In this paper, we first prove a lower bound, linear in the size of the PC, on the number of samples needed for \textit{any learning algorithm} to guarantee bounded statistical distance (Hellinger, reverse Kullback-Leibler, or total variation) with high probability. Next, for learning the parameters of deterministic and decomposable PCs under known structure, we also show a matching upper bound on the number of samples in the case of the Hellinger distance and reverse KL divergence. Dropping determinism, we use a tournament-based approach to provide an upper bound, proving that decomposable PCs are PAC learnable. Lastly, removing the known-structure assumption, we provide an instance-specific statistical upper bound on the total variation distance using the empirical estimator.
\end{abstract}

\section{Introduction}
As generative models capable of compactly representing probability distributions, probabilistic circuits (PCs) serve as a unifying framework to study many families of tractable probabilistic models, as PCs imposing different structural constraints.
Enforcing structural properties on PCs guarantees exact and efficient inference: precisely, polynomial (often linear) time with respect to the size of the circuit. Thus, understanding the \textit{expressive efficiency} of PCs---the size needed to represent a target function---is essential. The expressive efficiency of PCs classes has been studied extensively in the case of exact representation, as well as, to a lesser degree, the approximate case~\citep{darwicheACs,bova2016knowledge,choi2020probabilistic, leland2026hardness, martens2014expressive}.
Nevertheless, these works 
do not address the more common setting for PCs nowadays in practice: learning from data. That is, how does the number of samples needed to learn a PC relate to their size? 

We turn to probably approximately correct (PAC) learning theory to answer this question. PAC learning algorithms are analyzed with two central concerns: (1) \textit{time complexity} and (2) \textit{sample complexity} \citep{servedio2025probably}. Sample complexity characterizes the number of samples a learning algorithm requires to guarantee that the function it returns approximates the true function within a small error with high probability. Note that sample complexity alone \textit{does not} consider the runtime of a given algorithm. A class is considered to be \textit{efficiently PAC Learnable} if there exists a learning algorithm that is polynomial in both time and sample complexity. Many existing works have studied sample complexity and PAC density estimation for probabilistic models such as Bayesian Networks~\citep{dasgupta_sample_nodate, friedman1996sample} and Gaussian mixture models~\citep{ashtiani2020near}; however, the only work on sample complexity for PCs, to the best of our knowledge, is an upper bound for tree-shaped sum-product~networks~\citep{aden-ali_sample_2020} .

We present novel PAC-style bounds for more general classes of PCs over Boolean variables. First, we prove a lower bound on the sample complexity for \textit{any learning algorithm} when the true distribution is a PC of size $S$ (Theorem~\ref{thm:lowerBoundPCs}, Section~\ref{sec:lowerbounds}). This result applies to any family of models that subsumes, or compactly represents deterministic PCs with a known structure, such as sum-product networks \citep{poon2011sum}, arithmetic circuits \citep{darwicheACs}, etc.
Next, we show that the parameters of deterministic and decomposable PCs are efficiently PAC learnable by introducing an efficient parameter learning algorithm for PCs with a matching sample complexity upper bound, achieving a tight upper and lower bound of $\Theta(\frac{S + \log(1/\delta)}{\epsilon})$ on the sample complexity for the KL divergence, and $\Theta(\frac{S + \log(1/\delta)}{\epsilon^2})$ for the Hellinger distance, under the realizability assumption that the true distribution is a PC with known structure (Theorem~\ref{thm:SampleComplexityPCs}, Section~\ref{sec:upperbounds}). 
Our arguments also provide looser lower and upper bounds of $\Omega(\frac{S + \log(1/\delta)}{\epsilon})$ and $O(\frac{S + \log(1/\delta)}{\epsilon^2})$, respectively, for the total variation distance. 
To better complete our understanding of PC parameter learning, we use a tournament-based approach to provide a sample complexity upper bound of $O(\frac{S\log(S) + S\log(1/\delta)}{\epsilon^4} + \frac{\log(1/\delta)}{\epsilon^2})$ for learning decomposable PC parameters for the Hellinger and total variation distance (Theorem~\ref{thm:upperbound-nondet}, Section~\ref{sec:upperbounds}). This bound proves that decomposable PCs are PAC learnable in the case of the Hellinger distance and total variation distance, though the algorithm is inherently inefficient. 
While the realizable setting allows us to prove tight bounds, the assumption does not generally hold when learning PCs from data. Thus, we also provide a high-probability upper bound on the total variation distance between the true data distribution and a learned circuit by using the empirical distribution as an intermediary (Theorem~\ref{thm:empiricaldistTriangle}, Section~\ref{sec:Relaxing}).  

\section{Preliminaries}\label{sec:background}
\subsection{Sample Complexity}
For a given distance measure $D$, 
we wish to characterize the number of samples from a distribution $P$ that is necessary and sufficient to learn $P$ within distance $\epsilon > 0$ with error probability $\delta \in (0,1]$: i.e., to ensure $\Pr[D({\hat{P}}\lvert \rvert {P}) \leq \epsilon] \geq 1-\delta$. 
There are two common settings considered in PAC learning: \textit{realizable} and \textit{agnostic}. 
In the realizable case, we assume that the true distribution $P$ and the distribution $\hat{P}$ output by our learning algorithm are members of the same class $\C$. In the agnostic case, it is not necessary that $P \in \C$, and the objective becomes $\Pr[D({\hat{P}}\lvert \rvert{P}) \leq  \inf_{\hat{P} \in \C} D({\hat{P}}\lvert \rvert{P})+\epsilon] > 1-\delta$; see that the error also depends \textit{structurally} on the choice of hypothesis class $\C$.
Informally, \emph{upper bounds} on the sample complexity state that \textit{there exists} an algorithm that, given at least $M$ samples from $P$, outputs $\hat{P}$ such that $\Pr[D({\hat{P}}\lvert \rvert{P}) \leq  \epsilon] > 1-\delta$; on the other hand, a \emph{lower bound} states that \textit{any algorithm} must have at least $M$ samples for the same guarantee.
For the distance measure $D$, we consider the total variation distance, reverse Kullback-Leibler (KL) divergence, and Hellinger distance, defined between discrete distributions as follows, respectively:
    \begin{gather*}
        \Dtv{\hat{P}}{P} = \frac{1}{2}\sum_{\x} \abs{\hat{P}(\x)-P(\x)}, \\
        \Dkl{\hat{P}}{P} = \sum_{\x} \hat{P}(\x)\log\bigg(\frac{\hat{P}(\x)}{P(\x)}\bigg), \\
        \Dh{\hat{P}}{P} = \sqrt{1 - \sum_{\x} \sqrt{\hat{P}(\x)P(\x)}}.
    \end{gather*}
We use the \textit{squared} Hellinger for the majority of our proofs due to its simple formulation and tensorization, but the bounds trivially extend to the Hellinger distance. We also use the following relationships between these measures:
\begin{lemma}
    [Divergence Inequalities \citep{canonne2020survey}]\label{lemma:inequalities}
    The following hold:
    (1) $\Dtv{\hat{P}}{P} \!\leq\! \sqrt{\frac{1}{2}\Dkl{\hat{P}}{P}}$;
        (2) $\Dh{\hat{P}}{P}^2 \!<\! \Dtv{\hat{P}}{P}$;
        (3)$\Dh{\hat{P}}{P}^2 \!<\! \frac{1}{2}\Dkl{\hat{P}}{P}.
    $
\end{lemma}

\subsection{Probabilistic Circuits}
\begin{figure}[t]
\centering
\scalebox{0.55}{
    \begin{tikzpicture}[every node/.style={inner sep=0pt, minimum size=8pt}]
    
      \sumnode[line width=\midlinewidth]{s1};
    
      \prodnode[line width=\midlinewidth, below left=0.9cm of s1, xshift=-25pt]{p21};
      \prodnode[line width=\midlinewidth, below right=0.9cm of s1, xshift=+25pt]{p22};
      \weigedge[line width=\midlinewidth,left=0.5cm of s1,pos=0.3] {s1} {p21} {\Large$\theta_{1}$};
      \weigedge[line width=\midlinewidth,right=0.5cm of s1,pos=0.3] {s1} {p22} {\Large$\theta_{2}$};
    
      \sumnode[line width=\midlinewidth, below=0.6cm of p21]{s31};
      \bernode[line width=\midlinewidth, below left=0.6cm of p21, xshift=-5pt]{v31}{\large$X_3$};
      \sumnode[line width=\midlinewidth, below=0.6cm of p22]{s32};
      \bernode[line width=\midlinewidth, below right=0.6cm of p22, xshift=5pt]{v32}{\large$\neg X_3$};
      \edge[line width=\midlinewidth]{p21}{s31,v31};
      \edge[line width=\midlinewidth]{p22}{s32,v32};
      
      \prodnode[line width=\midlinewidth,below=0.6cm of s31,xshift=-0.6cm]{p41};
      \prodnode[line width=\midlinewidth,below=0.6cm of s31,xshift=0.6cm]{p42};
      \prodnode[line width=\midlinewidth,below=0.6cm of s32,xshift=-0.6cm]{p43};
      \prodnode[line width=\midlinewidth,below=0.6cm of s32,xshift=0.6cm]{p44};
      
      \edge[line width=\midlinewidth]{s31}{p41};
      \edge[line width=\midlinewidth]{s31}{p42};
      \edge[line width=\midlinewidth]{s32}{p43};
      \edge[line width=\midlinewidth]{s32}{p44};
    
      \sumnode[line width=\midlinewidth, below=0.5cm of p41]{s51};
      \sumnode[line width=\midlinewidth, below=0.5cm of p42]{s52};
      \sumnode[line width=\midlinewidth, below=0.5cm of p43]{s53};
      \sumnode[line width=\midlinewidth, below=0.5cm of p44]{s54};
      
      \bernode[line width=\midlinewidth,below left=0.5cm of p41, xshift=-5pt]{v51}{\large$X_2$};
      \bernode[line width=\midlinewidth,below right=0.5cm of p42, xshift=5pt]{v52}{\large$\neg X_2$};
      \bernode[line width=\midlinewidth,below left=0.5cm of p43, xshift=-5pt]{v53}{\large$X_2$};
      \bernode[line width=\midlinewidth,below right=0.5cm of p44, xshift=5pt]{v54}{\large$\neg X_2$};
    
      \edge[line width=\midlinewidth]{p41}{v51,s51};
      \edge[line width=\midlinewidth]{p42}{v52,s52};
      \edge[line width=\midlinewidth]{p43}{v53,s53};
      \edge[line width=\midlinewidth]{p44}{v54,s54};
    
      \bernode[line width=\midlinewidth,below=0.5cm of s51]{v61}{\large$X_1$};
      \bernode[line width=\midlinewidth,below=0.5cm of s52]{v62}{\large$\neg X_1$};
      \bernode[line width=\midlinewidth,below=0.5cm of s53]{v63}{\large$X_1$};
      \bernode[line width=\midlinewidth,below=0.5cm of s54]{v64}{\large$\neg X_1$};
    
      \edge[line width=\midlinewidth]{s51}{v61,v62};
      \edge[line width=\midlinewidth]{s52}{v61,v62};
      \edge[line width=\midlinewidth]{s53}{v63,v64};
      \edge[line width=\midlinewidth]{s54}{v63,v64};

    \end{tikzpicture}}
\caption{A smooth, decomposable, and deterministic PC (weights shown only at the root for brevity).}
\label{fig:ourcircuit}
\end{figure}
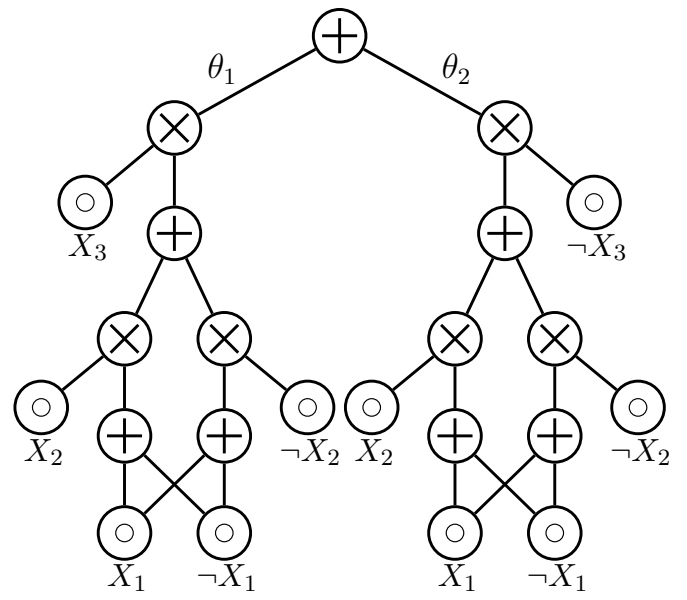
Probabilistic circuits (PCs) \citep{choi2020probabilistic} provide a unifying framework for many tractable probabilistic models, such as arithmetic circuits \citep{darwiche2002logical}, sum-product networks \citep{poon2011sum}, cutset networks \citep{rahman2014cutset}, probabilistic sentential decision diagrams \citep{kisa2014probabilistic}, and bounded-treewidth graphical models \citep{elidan2008learning,ChowLiu}. 

\begin{definition}[Probabilistic circuits] A probabilistic circuit (PC) \(\mathcal{C} := (\mathcal{G},\theta) \) represents a joint probability distribution \(P(\X)\) over random variables \(\X\) through a directed acyclic graph (DAG) \(\mathcal{G}\) parameterized by \(\theta\). The DAG is composed of 3 types of nodes: leaf, product, and sum nodes. Every leaf node $n$ in \(\mathcal{G}\) is an input function over a variable $X\in\X$, also called its scope $\phi(n)$.
Every internal node $n$ receives inputs from its children \(\ch(n)\), and its scope is defined recursively as $\phi(n) = \cup_{c \in \ch(n)} \phi(c)$.
Each node $n$ of a PC recursively defines a distribution as follows:
\begin{equation*}\label{eq:PC}
    P_n(\x) := \begin{cases}
        l(\x), & \text{ if $n$ is a leaf,}                         \\
        \prod_{c\in \ch(n)} P_c(\x) & \text{ if $n$ is a product,}           \\
        \sum_{c\in \ch(n)} \theta_{n,c} P_c(\x)  & \text{ if $n$ is a sum,} \\
    \end{cases}
\end{equation*}
where \(\theta_{n,c} \in [0,1]\) is the parameter associated with the edge connecting nodes \(n,c\) in \(\mathcal{G}\), and \(\sum_{c\in \ch(n)} \theta_{n,c} = 1.\) 
The distribution represented by the circuit is the output at its root node.
In this paper, we assume that $l(x)$ is a Boolean indicator function: i.e., $\mathbb{I}[x=1]$ or $\mathbb{I}[x=0]$, and the size of a PC (denoted $S$) refers to the number of parameters.
\end{definition}

Probabilistic circuits guarantee tractable inference
by composing sum and product gates under specific conditions such as \textit{smoothness}, \textit{decomposability}, and \textit{determinism} (see Figure~\ref{fig:ourcircuit} for an example). 
As is common in the literature, we assume that PCs are smooth and decomposable, which enable tractable marginal and conditional inference as well as sampling.
\begin{definition}[Smoothness and decomposability]\label{def:sm}
A PC is \textit{smooth} if the children of every sum node $n$ have identical scopes: \(\phi(c) = \phi(n),\, \forall c \in \ch(n)\). It is \textit{decomposable} if the children of every product node $n$ have disjoint scopes: \(\phi(c_i) \cap \phi(c_j) = \emptyset,\, \forall c_i \neq c_j \in \ch(n)\).
\end{definition}

\begin{definition}[Determinism]\label{def:det}
A PC is \textit{deterministic} if for every sum node and any fully-instantiated input, the output of at most one of its children is nonzero; i.e., the supports of its children are mutually disjoint.
\end{definition}

Learning the parameters of a deterministic PC with the closed-form MLE algorithm can be viewed intuitively as learning many categorical distributions using the empirical (counting) estimator, where each categorical variable corresponds to a sum node in the circuit with as many categories as there are children. This can be formalized using the notion of \textit{contexts}~\citep{liang2017learning,ChoiAAAI21}.
\begin{definition}
    [Context]\label{def:context}
    The \textit{context} $\gamma_n$ of every node $n$ in a PC, is defined in a top-down manner: for the base case, the context of the root node $n_r$ is the support $\gamma_{n_r} = \supp(n_r)$. For all other nodes $n$, its context is the intersection of its support and the union of its parents' context: \[\gamma_n = \bigcup_{m \in \mathsf{pa}(n)} \gamma_m \cap \supp(n).\]
\end{definition}
Given a sample $\x$, the value of the categorical variable corresponding to a sum node $n$ is given only if $\x\in\gamma_n$, with the specific value being the child index $i$ such that $\x\in\gamma_{n_i}$ as well; this can be evaluated for all sum nodes by propagating the sample through the circuit. We denote the probability of reaching a node $n$ as $P(\gamma_n)$. 

Determinism provides further desirable properties. Namely, given a joint assignment $\x$, a deterministic PC induces \textit{exactly one} tree-shaped sub-circuit through which $P(\x)$ is evaluated~\citep{choi2017relaxing}. Then, computing the probability of $\x$ for a deterministic PC amounts to simply multiplying the parameters associated with the induced sub-tree for $\x$. Formally, we denote this tree $\tree(\x)$, and thus for a deterministic circuit we write $P(\x) = \prod_{e \in \tree(\x)} \theta_e$. This is especially of use when we define both circuits to have the same underlying graph, as a sample must follow identical paths through the circuit as the underlying structure and leaves are the same.

\section{Lower Bounds for Learning PCs}\label{sec:lowerbounds}
This section investigates the following question: \textit{how many samples are necessary to ensure that we can learn a PC within a bounded distance to the true distribution with high probability?} While this is a natural question in learning theory, this has not been answered explicitly for any hypothesis class of PCs.
In particular, we consider the classes of deterministic PCs of fixed structure
and prove that the required sample complexity is polynomial in circuit size $S$, accuracy parameter $1/\epsilon$, and confidence parameter $1/\delta$.
\begin{theorem}
    [Lower Bound]\label{thm:lowerBoundPCs}
    Let $\C_{S}$ be a family of distributions represented by PCs with $S$ parameters.
    For $0 < \epsilon = O\left(\sqrt{\min_n P(\gamma_n)}\right) $, $\delta \in (0,0.15]$, any learning algorithm $A$ that takes in $M$ samples from $P\in \C_S$ and outputs $\hat{P}$ such that $\Dh{\hat{P}}{P} \leq \epsilon$ with probability $> 1-\delta$ requires 
    \begin{align*}
        M \geq \max(\frac{567S}{131072\epsilon^2},\frac{\log(2/\delta)}{5\epsilon^2}) = \Omega(\frac{S + \log(1/\delta)}{\epsilon^2}),
    \end{align*}
    $\Dkl{\hat{P}}{P} \leq \epsilon$ with prob. $1-\delta$ requires
    \begin{align*}
        M \geq \max(\frac{567S}{65536\epsilon},\frac{2\log(2/\delta)}{5\epsilon}) = \Omega(\frac{S + \log(1/\delta)}{\epsilon}),
    \end{align*}
    and $\Dtv{\hat{P}}{P} \leq \epsilon$ with prob. $1-\delta$ requires,
    \begin{align*}
        M \geq \max(\frac{567S}{131072\epsilon},\frac{\log(2/\delta)}{5\epsilon^2}) = \Omega(\frac{S + \log(1/\delta)}{\epsilon}),
    \end{align*}
    even if $P$ is in $C_{\mathcal{G}}$, a family of deterministic PCs of a fixed structure $\mathcal{G}$ with $S$ parameters, and $\mathcal{G}$ is known to $A$.
\end{theorem}

Notably, this result \textit{does not restrict} the output of learner $A$, and applies to \textit{any} PC class that subsumes deterministic PCs of any fixed structure(s). 

We first give a high-level proof sketch here before presenting the formal proof.
The proof relies on standard lower-bound techniques for discrete distributions: Assouad's Lemma and the hardness of estimating the bias of a coin with high probability.
Assouad's Lemma provides a guiding principle to constructing a family of distributions which requires many samples to learn. These distributions are constructed by making many different ``choices'', by perturbing an initial distribution slightly with each choice; for instance, we will construct such family of PCs by perturbing the parameters, by a carefully chosen amount, around the uniform distribution (Proposition~\ref{prop:construction}).
This family will include a large set of distributions that are difficult to tell apart by any learning algorithm without using many samples. 
Moreover, suppose we have two distributions in this family that differ by only one perturbation; this perturbation can be mapped to the task of biasing a fair coin. As such, if an algorithm can learn the constructed family of distributions with high probability, then it also learns the bias of a coin with high probability, thereby inheriting its sample complexity. 
\begin{lemma}
    [Assouad's Lemma~\citep{assouad1983deux,canonne2020survey}]\label{lem:Assouad}
    Let $\C$ be a family of probability distributions and $A$ be the set of learning algorithms that take $M$ samples and output a distribution $\hat{P}$. Suppose there exists a family of $\br{P_v}_{v \in \br{-1,1}^r} \subseteq \C$ of $2^r$ distributions and constants $\alpha,\beta > 0$ such that $\forall\, v,v' \in \br{-1,1}^r$: 
    \begin{enumerate}[noitemsep]
        \item The distance between $P_v,P_{v'}$ is at least proportional to the Hamming distance $H$: \(\Dh{P_v}{P_{v'}}^2 \geq \alpha H(v,v')\);
        \item If $ H(v,v') = 1$, then \(\Dh{P_v}{P_{v'}}^2 \leq \beta.\)
    \end{enumerate}
    Then, for all $M \geq 1$, 
    \begin{equation*}
        \inf_{\hat{P}\in A} \sup_{P \in \C} \E_{s_1, \ldots, s_M}[\Dh{\hat{P}}{P}^2] \geq \frac{\alpha}{8}r(1-\beta)^{2M}.
    \end{equation*}
\end{lemma}

\begin{proofsketch}
    The original statement (seen in \citep{tsybakovNonparameteric}) is proven by observing that the average is smaller than the maximum and using LeCam's method \citep{yu1997assouad}. Lemma~\ref{lem:Assouad} differs from \citep{canonne2020survey} as we are using the squared Hellinger distance as our loss function. Thus, there is an additional factor of $1/2$ to account for the fact that the squared Hellinger only satisfies the weak triangle inequality. For a detailed explanation connecting the version found in \citep{tsybakovNonparameteric} to Lemma~\ref{lem:Assouad}, see the Appendix.
\end{proofsketch}
Informally, Assouad's Lemma interprets the distributions in the constructed family as being determined by $r$ distinct choices (or bit flips), and states that any learning algorithm must learn many choices when the Hamming distance $H(v,v')$ is large and simultaneously cannot easily distinguish two distributions if $H(v,v') = 1$~\citep{canonne2020survey}.
Next, we describe our constructed family of PCs that satisfies the conditions in Assouad's Lemma.
\begin{proposition}
    [PC Family Construction]\label{prop:construction}
    There exists a family of $2^{\lceil S/2 \rceil}$ distributions in $\C_\mathcal{G} \subseteq \C_S$ for a deterministic PC structure $\mathcal{G}$ of size $S$,  
    such that if $\eta \leq \min\frac{P(\gamma_n)}{16\abs{\ch(n)}D^2}$, $\forall v,v'\in \{-1,1\}^{S/2}$, \(\Dh{P_v}{P_{v'}}^2 \geq \frac{189}{256}\eta H(v, v')\) and if $H(v,v') \!=\! 1$, \(\Dh{P_v}{P_{v'}}^2 \leq \frac{4}{3}\eta.\)
\end{proposition}
\begin{proofsketch}
    W.l.o.g.\ suppose every sum node has an even number of children, and assign  an index in $[1,S/2]$ to each pair of edges of the same sum node. Then, define $P_v$ for each $v\in \{-1,1\}^{S/2}$ such that $\theta_{n,c}[v_{n,c}] = \frac{1}{\ch(n)} + v_c\tau_n$ 
    where $\tau_n = \sqrt{\frac{\eta}{\abs{\ch(n)}P(\gamma_n)} - \frac{\eta^2}{4P(\gamma_n)^2}}$. Using the conditions on $\eta$ gives the desired result.
\end{proofsketch}
Intuitively, using Assouad's Lemma on a smaller subset of larger family proves lower bounds on the larger family for two reasons: (1) Assouad's Lemma holds for \textit{any learning algorithm}, and (2) sample complexity considers all possible distributions, thus a lower bound on the worst subset is also a lower bound on the sample complexity of the larger family. We next obtain our lower bound from Assouad's Lemma with this construction and using the hardness of learning the bias of~a~coin with high probability.
\begin{proof}
    [Proof of Theorem~\ref{thm:lowerBoundPCs}]
We first apply Lemma~\ref{lem:Assouad} using the family of distributions constructed in Proposition~\ref{prop:construction} to obtain:
\begin{align*}
    &\inf_{A \in {A}_M} \sup_{P \in \C_S}\E_{s_1, \ldots, s_M \sim P}[\Dh{\hat{P}}{P}^2]\\
    &\geq \inf_{A \in {A}_M} \sup_{P \in \C_{\mathcal{G}}}\E_{s_1, \ldots, s_M \sim P}[\Dh{\hat{P}}{P}^2]\\
    &\geq \frac{1}{8}\alpha\frac{S}{2}(1-\beta)^{2M}
    \geq \frac{1}{8}\frac{189\eta}{256}\frac{S}{2}(1-\frac{4}{3}\eta)^{2M}\\
    &= \frac{189\eta S}{4096}(1-\frac{4}{3}\eta)^{2M}.
\end{align*}
To make $\frac{189\eta S}{2048} = 2\epsilon^2$, set $\epsilon =  O\left(\sqrt{\min_n P(\gamma_n)}\right)$, then $\eta = \frac{8192\epsilon^2}{1089S}$ which implies that 
$2\epsilon^2(1-\frac{32768\epsilon^2}{567S})^{2M}$.
Thus we must set, $(1-\frac{32768\epsilon^2}{567S})^{2M} < 1/2$ to guarantee $2\epsilon^2(1-\frac{32768\epsilon^2}{567S})^{2M} \leq \epsilon^2$. Using Bernoulli,
\begin{align*}
     1/2 > \left(1 - \frac{4}{3}\eta\right)^{2M}
     &\geq \left(1 - \frac{8}{3}\eta M\right)
\end{align*}
Therefore, $M \geq \frac{3}{16\eta} = \frac{567S}{131072\epsilon^2}$. Next we convert this expectation into a fixed probability statement. From our construction, 
\[
    \Dh{P_{\hat{v}}}{P_v}^2 \leq \frac{4}{3}\eta\frac{S}{2} \leq \frac{16384}{567}\epsilon^2.
\]
The reverse Markov inequality gives us:
\begin{align*}
   \Pr\left[\Dh{P_{\hat{v}}}{P_v}^2 \geq \frac{\epsilon^2}{2}\right] &\geq \frac{\epsilon^2-\epsilon^2/2}{\frac{16384}{567}\epsilon^2 - \epsilon^2/2}
   = \frac{567}{32201}.
\end{align*}
By the weak triangle inequality, $\Dh{\hat{P}}{P_v}^2\geq \frac{1}{4}\Dh{P_{\hat{v}}}{P_v}^2$, which implies finally that if $M\leq \frac{567S}{131072\epsilon^2}$ then $\sup_v \Pr\left[\Dh{\hat{P}}{P_v} \geq \frac{\epsilon}{\sqrt{8}}\right] \geq \frac{567}{32201}$. This yields the $M = \Omega\left(\frac{S}{\epsilon^2}\right)$ term.

See that the above only applies to the fixed probability regime. To move to any $\delta$, see that using our construction any learning algorithm must both be able to learn the above distribution \emph{and} be able to learn the bias of a coin (flip two edges for one sum node in our construction). Learning the bias of a coin is a well studied task, and is known to require at least $M = \frac{\log(2/\delta)}{5\epsilon^2}  = \Omega(\frac{\log(1/\delta)}{\epsilon^2})$ samples to learn within probability $1-\delta$.\footnote{For a simple proof of the upper and lower bounds, see \citet{dasgupta_sample_nodate}} 
Combining the two complexity results and taking the Hellinger distance as the distance measure, since $\Dh{\hat{P}}{P}^2 \geq \epsilon^2 \implies \Dh{\hat{P}}{P} \geq \epsilon$, we have that 
ensuring $\Dh{\hat{P}}{P} \leq \epsilon$ with probability $> 1-\delta$ requires:
\begin{align*}
    M \geq \max\left(\frac{567S}{131072\epsilon^2},\frac{\log(2/\delta)}{5\epsilon^2}\right),
\end{align*}
Similarly, using $\Dh{\hat{P}}{P}^2 < \frac{1}{2}\Dkl{\hat{P}}{P}$, ensuring $\Dkl{\hat{P}}{P} \leq \epsilon$ with probability $1-\delta$ requires:
\begin{align*}
    M \geq \max\left(\frac{567S}{65536\epsilon},\frac{\log(2/\delta)}{5\epsilon}\right),
\end{align*}
and finally since $\Dh{\hat{P}}{P}^2  \geq \epsilon \implies \Dtv{\hat{P}}{P} \geq \epsilon$, ensuring $\Dtv{\hat{P}}{P} \leq \epsilon$ with probability $1-\delta$ requires:
\begin{align*}
    M \geq \max\left(\frac{567S}{131072\epsilon},\frac{\log(2/\delta)}{5\epsilon^2}\right).
\end{align*}
\end{proof}
Notice in the proof above that we must scale $\epsilon$ with $\sqrt{\min P(\gamma_n)}$. Of course, for some structures such as shallow mixtures (i.e.\ single sum node) this is equivalent to requiring $\epsilon < 1$; thus there exist structures where this statement is trivial. However, there may also exist structures where without setting $\epsilon$ to be quite small, nodes deep in the circuit do not contribute enough to the sample complexity bound. This motivates further study into pruning circuit nodes, as discussed by \citet{dang2022sparse}. In short, we have shown that a learning algorithm for \textit{any} family of models that subsumes, or can compactly represent, deterministic PCs of fixed structure requires at least proportionally as many samples as the circuit size. For instance, this lower bound immediately applies to learning sum-product networks~\citep{poon2011sum}, arithmetic circuits~\citep{darwicheACs}, sum-of-squares circuits~\citep{loconte2025sum}, InceptionPCs~\citep{wang2025relationship}, probabilistic generating circuits~\citep{zhang2021probabilistic}, and more.

\section{Tight Sample Complexity Bounds}\label{sec:upperbounds}
We next show that this \textit{necessary} number of samples is sufficient to learn the parameters of a given deterministic PC structure to  guarantee KL divergence and Hellinger distance less than $\epsilon$ with probability $1-\delta$. This proves that the parameters of deterministic and decomposable PCs are \textit{efficiently} PAC learnable. We also prove a polynomial upper bound on the sample complexity of learning non-deterministic circuits; hence proving that their parameters are also PAC learnable, although not using an efficient (polytime) algorithm.

\subsection{Deterministic PCs}
\begin{theorem}
    [Upper Bound]\label{thm:SampleComplexityPCs}
    Let $\C_{\mathcal{G}}$ be a family of distributions represented by deterministic PCs of a fixed structure $\mathcal{G}$ of size $S$. 
    For $\epsilon > 0$ and $\delta \in (0,1]$, the closed-form estimator given $M$ samples from $P\in \C_{\mathcal{G}}$ outputs $\hat{P}$ such that $\Dkl{\hat{P}}{P} \leq \epsilon$ with probability $>\! 1\!-\!\delta$~if:
    \begin{align*}
        M \geq \max(\frac{15S}{e\epsilon}, \frac{2\log(1/\delta)}{\epsilon}) = O(\frac{S + \log(1/\delta)}{\epsilon}),
    \end{align*}
    and $\Dh{\hat{P}}{P} \!\leq\! \epsilon$ and $\Dtv{\hat{P}}{P} \!\leq\! \epsilon$ with prob.\ $>\! 1\!-\!\delta$ if:
    \begin{align*}
        M \geq \max(\frac{15S}{e\epsilon^2}, \frac{2\log(1/\delta)}{\epsilon^2}) = O(\frac{S + \log(1/\delta)}{\epsilon^2}).
    \end{align*}    
\end{theorem}

With our intuition for learning a deterministic PCs parameters with the closed form MLE, it is tempting to assume that the sample complexity can be derived trivially by upper bounding the sample complexity of learning a discrete distribution for each sum node. However, unlike learning discrete distributions where every sample corresponds to some category, in deterministic PCs not every sample is routed through every sum node, and a given sample can be completely irrelevant to many sum nodes. That is, from the total number of samples used to learn the parameters of a PC, only a fraction would be relevant for learning the parameters of each sum node $n$, directly related to the magnitude of $P(\gamma_n)$. Therefore, we must move past this naive approach.

The proof of Theorem~\ref{thm:SampleComplexityPCs} proceeds as follows: First, we prove a concentration inequality for the reverse KL divergence between deterministic PCs, using the fact the parameters of each sum node are learned using the empirical estimator. Then, we follow the argument from \citet{canonne2020short} in conjunction with our concentration inequality to yield the desired result. We highlight again that we use the \textit{{reverse KL divergence}}, which is necessary as enforcing a learning guarantee less than $\epsilon$ with the standard KL divergence requires learning arbitrarily small probabilities without approximating by zero~\citep{canonne2020short}. The reverse KL divergence between deterministic PCs has the following property,
\begin{lemma}
    [Deterministic reverse KLD Decomposition]\label{lemma:reverseKLPC}
    Let $\hat{P}$ and $P$ be two distributions over variables $\Y$ represented by deterministic PCs with the same structure. Then:
    \[\KL{\hat{P}(\Y)}{P(\Y)} = \sum_{\text{sum node } n} \hat{P}(\gamma_n)\KL{\hat{\theta}_{n}}{\theta_{n}},
    \] where $\hat{P}(\gamma_n)$ is the probability of reaching sum node $n$ under the learned circuit, and $\KL{\hat{\theta}_{n}}{\theta_{n}}$ is the reverse KL divergence of the categoricals associated with node $n$.
\end{lemma}

\begin{proofsketch}
    For the full proof see the Appendix. First, use the fact that as $\hat{P}$ and $P$ are deterministic and thus evaluating a given $\y$ creates a tree through the circuit. This allows us to turn $\log(\frac{\hat{P}(\y)}{P(\y)})$ into $\sum_{e \in \tree(\y)} \log(\frac{\hat{\theta}_e}{\theta_e})$. Using indicator functions we can change this to be sums over the nodes and children. Using this in the definition of the reverse KL divergence yields the result.
\end{proofsketch}

This result essentially follows from the fact that circuits with the same structure and leaves only differ in parameterization. Determinism simply ensures these differences can be taken locally at each sum node and combined by weighting with the probability of reaching a given sum node.
\begin{theorem}\label{thm:concentrationPC}
For $P$ a deterministic PC and $\hat{P}$ learned with MLE with the same underlying structure and $S$ parameters, and for all $\epsilon > \frac{S}{M}$,
    \[\Pr[\Dkl{\hat{P}}{P} \geq \epsilon] \leq e^{-\epsilon M}\bigg(\frac{\epsilon e M}{{S}}\bigg)^{{S}}.\]
\end{theorem}

Which relies on the following result from \citet{agrawal2020finite},
\begin{theorem}[\citet{agrawal2020finite}]\label{thm:MGFUpper}
    For $P$ a categorical distribution over $k$ categories and $\hat{P}$ the empirical estimator with $n$ samples, it holds that for all $0\leq t < n$,
    \[
        \E[\exp(t\Dkl{\hat{P}}{P})] \leq \bigg(\frac{1}{1-t/n}\bigg)^{k-1}.
    \]
\end{theorem}
Given that we are using the closed form MLE, which is equivalent to the empirical estimator described by \citet{agrawal2020finite}, if each sum node receives $m_n$ samples we can transfer the upper bound from $\E[\exp(t\Dkl{\hat{\theta}_{n,c}}{\theta_{n,c}})]$. However, in PCs, the number of samples to reach node $n$ is determined by a random variable, $M_n$. Notice that the number of samples to reach node $n$ also depends on the sum of the number of samples to reach the parents of $n$ (at the root this is $M$), thus there exists a dependence between sum nodes. To circumvent this, we use the law of iterated expectation topologically down the circuit, and upper bound each sum node as we traverse downwards on the circuit using Theorem~\ref{thm:MGFUpper}.
\begin{proof}
    [Proof of Theorem~\ref{thm:concentrationPC}]
    Let $\mathcal{M}$ be the set of all random variables $M_n$. Using Lemma~\ref{lemma:reverseKLPC}, 
    \begin{align*}
        \E[\exp(t\Dkl{\hat{P}}{P})] = \E[\exp(t \sum_n \hat{P}(\gamma_n)\Dkl{\hat{P}_{n,c}}{P_{n,c}})].
    \end{align*} 
    Now, we order all sum nodes topologically from parents to children, and define $F_{<n}$ to record the routing of samples preceding node $n$. Thus for every node $n$ we know $M_n = m_n$ and the choices among the children of $n$ are independent. See first that upon fixing $M_n$ the closed-form MLE yields $\hat{P}(\gamma_n) = \frac{m_n}{M}$. At each sum node then we use Theorem~\ref{thm:MGFUpper} conditionally to get
    \[
        \E[\exp(t\frac{m_n}{M}\Dkl{\hat{\theta}_n}{\theta_n}) \vert F_{<n}] \leq  \bigg(\frac{1}{1-\frac{t}{M}}\bigg)^{\abs{\ch(n)}-1}.
    \]
    Now, using the law of iterated expectation topologically, 
    \begin{align*}
        &\E[\exp(t\Dkl{\hat{P}}{P})]\\
        &\leq \E[ \exp(t \sum_{b < n} \frac{m_b}{M}\Dkl{\hat{\theta}_b}{\theta_b})\cdot\\
        & \qquad \qquad \qquad \E[\exp(t\frac{m_n}{M}\Dkl{\hat{\theta}_n}{\theta_n}) \vert F_{<n}]]\\
        &\leq \bigg(\frac{1}{1-\frac{t}{M}}\bigg)^{\abs{\ch(n)}-1} \E[ \exp(t \sum_{b < n} \frac{m_b}{M}\Dkl{\hat{\theta}_b}{\theta_b})]
    \end{align*}
    Iterating over all nodes we get: 
    \begin{align*}
        \E[\exp(t\Dkl{\hat{P}}{P})] &\leq \prod_n  \bigg(\frac{1}{1-\frac{t}{M}}\bigg)^{\abs{\ch(n)}-1}\\
        &\leq \bigg(\frac{1}{1-\frac{t}{M}}\bigg)^{S}.
    \end{align*}
    Using a Chernoff bound, for $0 \leq t < M$ we have
    \begin{align*}
        \Pr[\Dkl{\hat{P}}{P} \geq \epsilon] 
        &\leq \inf_{t \in [0,M)} e^{-t\epsilon}\bigg(\frac{1}{1-\frac{t}{M}}\bigg)^{S},
    \end{align*}
    Choose $\frac{t}{M} = 1 - \frac{S}{\epsilon M}$, then for all $\epsilon > \frac{S}{M}$,
    \begin{align*}
        \Pr[\Dkl{\hat{P}}{P} \geq \epsilon]
        \leq e^{-\epsilon M}\bigg(\frac{\epsilon e M}{{S}}\bigg)^{{S}}.
    \end{align*} 
\end{proof} 

At last, using Theorem~\ref{thm:concentrationPC} we are ready to prove Theorem~\ref{thm:SampleComplexityPCs} following a similar argument from \citet{canonne2020short}.
\begin{proof}
    [Proof of Theorem~\ref{thm:SampleComplexityPCs}]
    Let $\hat{P}$ and $P$ be over $\X$ where $P$ and $\hat{P}$ have the exact same underlying graph. Then, using the change of exponent rules, we rewrite Theorem~\ref{thm:concentrationPC} as
    \begin{align*}
        \Pr[\Dkl{\hat{P}(\X)}{P(\X)} \geq \epsilon] &\leq e^{{S}\log(\frac{\epsilon e M}{{S}})}e^{-\epsilon M}\\
        &= e^{-\epsilon M + S\log(\frac{\epsilon e M}{S})}
    \end{align*}
    If $M \geq \frac{15{S}}{e\epsilon}$ then ${S}\log(\frac{e\epsilon M}{{S}}) < \frac{M\epsilon}{2}$. Thus, when $M \geq \frac{15{S}}{e\epsilon}$,
    \begin{align*}
        \Pr[\Dkl{\hat{P}(\X)}{P(\X)} \geq \epsilon] &\leq e^{-\frac{M\epsilon}{2}}.
    \end{align*}
    Finally, setting $e^{-\frac{M\epsilon}{2}} \leq \delta$ we get that $M \geq \frac{2\log(1/\delta)}{\epsilon}$ implies that $\Pr[\Dkl{\hat{P}(X)}{P(X)} \geq \epsilon] < \delta$. Thus, combining these two terms results in a guarantee learning of $\Dkl{\hat{P}(X)}{P(X)} \leq \epsilon$ with probability $1-\delta$ using at least $O(\frac{{S} + \log(1/\delta)}{\epsilon})$ samples. The extension to the KL divergence and squared Hellinger distance follows from Lemma~\ref{lemma:inequalities}, which results a learning guarantee for $\Dtv{\hat{P}}{P} \leq \epsilon$ and $\Dh{\hat{P}}{P} < \epsilon$ with probability $1-\delta$ using at least $O(\frac{{S} + \log(1/\delta)}{\epsilon^2})$ samples. 
\end{proof}

\subsection{Non-deterministic PCs}
We have shown a tight upper and lower bound on the number of samples needed for deterministic PCs of known structure. However, our lower bound holds for non-deterministic PCs as well. This leaves a clear question: \emph{what is the sample complexity of learning the parameters of \emph{non-deterministic PCs}?} Without determinism, we no longer have (1) closed-MLE nor (2) exactly one path through the circuit for any given sample, preventing us from naively applying our previous results; however, using the notion of a \textit{tournament},\footnote{Sometimes called a Scheff\'e tournament, see Chapter 6~\citep{devroye2001combinatorial}} we can provide an upper bound for learning non-deterministic PCs. 

\begin{theorem}
    [Sample Complexity of Learning Non-deterministic PCs]\label{thm:upperbound-nondet}
    Let $\C_{\mathcal{G}}$ be a family of distributions represented by non-deterministic PCs of a fixed structure $\mathcal{G}$ of size $S$. 
    For $\epsilon > 0$ and $\delta \in (0,1]$, there exists a learning algorithm such that given $M$ samples from $P\in \C_{\mathcal{G}}$ there exists $\hat{P}$ such that $\Dh{\hat{P}}{P} \leq \epsilon$ (or $\Dtv{\hat{P}}{P} \leq \epsilon$) with probability $>\! 1\!-\!\delta$~if
    \[M = O\left(\frac{S\log(S) + S\log(1/\delta)}{\epsilon^4} + \frac{\log(1/\delta)}{\epsilon^2}\right).\]
\end{theorem}
We first provide a sketch of our proof approach. Using  by the latent variable interpretation~\citep{peharz2016latent} every non-deterministic sum node has a latent variable $Z$ whose values are associated with each of its children. Then by making these latent variables explicit, which we call a ``latent-completion'', we can instantiate a deterministic and decomposable PC $P(\X,\Z)$ whose marginal probability distribution matches that given by the corresponding decomposable PC $P(\X)$.
Now, take $M$ samples $\x_1,\ldots, \x_M$ from the true distribution $P(\X)$. If we enumerate all possible latent variable assignments $\z$ for each sample to form the set of all possible completions $\{(\x_1,\z_1),\ldots, (\x_M,\z_M)\}$, this set \textit{must include} a dataset of $M$ samples from the true $P(\X,\Z)$. As such, for sample complexity of learning a circuit with the dataset from the true distribution, we can use the argument for deterministic PCs from Theorem~\ref{thm:SampleComplexityPCs}. However, we do not know exactly which latent-completion dataset matches the true distribution. To overcome this, we use a \textit{tournament} which is a common technique in learning theory when there exist many possible candidate distributions~\citep{canonne_testing_2020, daskalakis2012learning}. If there are at most $t$ candidate circuits, we run a pairwise tournament between all candidates, which requires $O(t^2)$ hypothesis tests; by choosing the candidate that wins the most rounds and taking a union bound, we are guaranteed to return a circuit whose distribution is $O(\epsilon)$ from the true distribution with high probability.

As our lower bound applies both the Hellinger distance and total variation distance, we use the tournament and test statistic described in \citep{suresh2021robust} to yield an upper bound for both. Unfortunately, in the current literature there is no tournament selection approach for the reverse KL-divergence, thus we are unable to provide an upper bound for decomposable PCs learned with the reverse KL divergence currently. Given $M$ samples $\X^{\otimes M}$ from an unknown distribution $P$ and two candidates $Q_1,Q_2$, suppose
\begin{align*}
H_0: & \quad \gamma \Dh{Q_2}{P} \leq \Dh{Q_1}{P},\\
    H_1: & \quad \Dh{Q_2}{P} \geq \gamma \Dh{Q_1}{P}.
\end{align*}
The test statistic for this hypothesis test is 
\begin{equation*}
    T(Q_1,Q_2,\X^{\otimes M}) = \frac{1}{M}\sum_{i=1}^M \frac{Q_1(\x^{(i)})-Q_2(\x^{(i)})}{Q_1(\x^{(i)}) + Q_2(\x^{(i)})},
\end{equation*}
where $\x^{(i)}$ is the $i^{\text{th}}$ sample and $Q_1$ wins if $T(Q_1,Q_2,\X^{\otimes M}) > 0$, $Q_2$ wins if $T(Q_1,Q_2,\X^{\otimes M}) < 0$, and the winner is picked randomly if $T(Q_1,Q_2,\X^{\otimes M}) = 0$. 
This procedure is referred to as the Hellinger Test; given at least $\frac{c_{\gamma} \log(1/\delta')}{\Dh{Q_1}{Q_2}^2} = O(\frac{\log(1/\delta')}{\Dh{Q_1}{Q_2}^2})$ samples for $\gamma > \frac{\sqrt{2}}{\sqrt{2}-1}$ and a constant $c_\gamma$ depending on $\gamma$ the Hellinger test ensures the selection of $Q_2$ under $H_0$ and $Q_1$ under $H_1$ with probability at least $1-\delta$~\citep{suresh2021robust}.

Using the Hellinger Test, choosing the candidate that wins the most matches guarantees the chosen distribution to be closest to the true distribution within a constant factor. 
\begin{lemma}\label{lem:winner}
Let $P$ be the true distribution and $T$ a set of candidate distributions. Suppose that $\min_{b \in T}\Dh{P}{b} \leq \frac{\epsilon}{\gamma^2}.$
Run the Hellinger test between every pairwise set of candidates in $T$ using 
$O(\frac{\log(1/\delta')}{\epsilon^2})$ samples, and let $f$ be the distribution that wins the most rounds of the Hellinger test. Then, 
\[
    \Pr[\Dh{P}{f} \leq \epsilon] \geq 1-\delta.
\]
\end{lemma}
\begin{proof}
    Suppose $f$ wins the most rounds and there exists $b \in T$ such that $\Dh{P}{b} \leq \epsilon/\gamma^2$. Suppose for contradiction that $\Dh{P}{f} > \epsilon  > \gamma^2 \Dh{P}{b}$. Define the set $A = \br{j: \gamma\Dh{P}{j} \leq \epsilon/\gamma}$; then for all $j \in A$, $\gamma\Dh{P}{j} \leq \Dh{P}{f}$ and for all $j \in A^c$, $\Dh{P}{j} > \gamma\Dh{P}{b}$. By the triangle inequality then, for $j \in A$, 
    $\Dh{f}{j} \geq \gamma\Dh{P}{j} - \Dh{P}{f} \geq \epsilon - \epsilon/\gamma$ and for $j \in A^c$, $\Dh{b}{j} \geq \gamma\Dh{P}{j} - \Dh{P}{b} \geq \epsilon/\gamma - \epsilon/\gamma^2$. Thus, as $\gamma$ is a constant, a union bound ensures all comparisons are correct with probability $1-\delta$ with at least $O(\frac{\log(\abs{T}) + \log(1/\delta)}{\epsilon^2})$ samples. 

    As we have confirmed our comparisons are correct, with probability $1-\delta$, $A$ is a subset of all losses for $f$. Following this $A^c \setminus \br{f}$ contains the set of all wins for $f$, and thus the number of wins for $f$ is at most $\abs{A^c}-1$. But, $A^c = \br{j: \Dh{P}{j} > \epsilon/\gamma}$, thus for every $j \in A^c$ candidate $b$ wins. As such the number of wins for $b$ is at least $\abs{A^c} > \abs{A^c} - 1$ which is greater than the number of wins for $f$. Therefore, $f$ cannot have the most wins which is a contradiction.
\end{proof}
Thus, our selection criteria will yield a candidate with high probability that is within at most a constant factor of any other. We next determine the number of possible latent completions. 
\begin{lemma}
    [Candidate Count]\label{lem:countingcompletions}
    Given $M$ independent samples from $P$, there are at most $S^M$ possible datasets $(\x,\z)$ from $P(\X,\Z)$ where the true distribution $P(\X)$ is a PC with $S$ parameters.
\end{lemma}
\begin{proof}
    Suppose we receive $M$ independent samples from $P$, that that each sum node has $c_s$ children, and a sample can be routed down each child. Knowing this, there are at most $c_s^M$ possibilities for any given sum node.
    Taking this across all sum nodes results in $S^M$ possible latent completions $(\X= \x,\Z = \z)$. 
\end{proof}
While method of counting possible latent-completions likely leads to over-counting this is acceptable as we are considering an upper bound. Finally, we have the ingredients needed, and prove Theorem~\ref{thm:upperbound-nondet}.
\begin{proof}[Proof of Theorem~\ref{thm:upperbound-nondet}]
    Let $P$ be the true distribution represented by a decomposable PC and $\hat{P}$ be a PC of the exact same structure with $S$ parameters. From \citep{peharz2016latent} the introduction of latent variables allows us to represent $P(\X)$ as $P(\X,\Z)$ which is a deterministic augmentation over $(\X,\Z)$. Let $\eta = \frac{\epsilon}{16}$ and draw a set of $M$ samples from $P$, called $\X^{\otimes M}$, where
    \[
        M = O\left( \frac{S + \log(1/\delta)}{\eta^2}\right).
    \]
    Now, if we are given both the samples $\X^{\otimes M}$ and the true latent completions $\Z$ from $P$ then learning $\hat{P}$ over $(\X,\Z)$ with the empirical estimator yields \( \Pr[\Dh{\hat{P}(\X,\Z)}{P(\X,\Z)} \leq \eta] \geq 1 - \delta \) by Theorem~\ref{thm:SampleComplexityPCs}. The monotonicity of $\Dh{\cdot}{\cdot}$ then proves \( \Pr[\Dh{\hat{P}(\X)}{P(\X)} \leq \eta] \geq 1 - \delta \). But we do not know exactly which dataset has the correct choice of $P(\Z\vert \X)$. Thus, we must enumerate all possible latent variable instantiations for a given dataset, call this larger set $L$. See that $(\X,\Z) \in L$ such that $(\X,\Z) \sim P(\X,\Z)$, and thus one of these datasets allows us to construct $\hat{P}$ from the above.

    Using Lemma~\ref{lem:countingcompletions}, we know that $\abs{L} \leq S^M$ and let $\gamma = 4 > \frac{\sqrt{2}}{\sqrt{2}-1}$. Conditionally on the number of samples already taken $M$, we run the Hellinger tournament on $L$. From above we know that $L$ contains a candidate within $\eta$ of $P$, and by Lemma~\ref{lem:winner} using a union bound over all tests in the tournament returns some $\hat{P}$ satisfying \(\Dh{\hat{P}}{P} \leq \epsilon\) with probability $1-\delta$
    given $O(\frac{\log(S^M) + \log(1/\delta)}{\epsilon^2}) = O(\frac{M\log(S) + \log(1/\delta)}{\epsilon^2})$ samples. 
    Substituting in our original number of samples $M$ gives us the desired result of 
    \begin{align*}
        O\left(\frac{S\log(S) + S\log(1/\delta)}{\epsilon^4} + \frac{\log(1/\delta)}{\epsilon^2}\right).
    \end{align*}
    The total variation distance follows from the same argument up to constant factors as $\Dtv{\hat{P}}{P} \leq \sqrt{2}\Dh{\hat{P}}{P}$.
\end{proof}

This proves that non-deterministic PCs are PAC learnable for the Hellinger and total variation distance, though not with an efficient algorithm due to the exponential enumeration required by the tournament. Interestingly, this shows that while non-deterministic PCs can be exponentially more efficient than deterministic PCs, they require only a polynomially greater number of samples.

\section{Relaxing Structural Assumptions}\label{sec:Relaxing}

The upper bounds discussed above are derived using the restrictive assumption that the true distribution can be represented as a PC with a known structure. However, in practice when learning PCs from data this assumptions is unlikely to be true. As a step towards establishing sample complexity in the agnostic setting, we provide a computable bound on the distance between a learned PC and an unknown distribution $P$, accessible only through a set of samples from it.

Let $\Tilde{P}$ be the empirical distribution induced by a dataset $\mathcal{D}$. We define the half-norm as follows: 
\begin{definition}
    [Half-Norm]$\|\Tilde{P}\|_{1/2} := \left( \sum_{i} \sqrt{\Tilde{P}_i}\right)^2$ where $\Tilde{P}_i = \frac{1}{m} \sum_{t \in [m]} \mathbb{I}(\X = i)$ is the probability of the $i^{\text{th}}$ category.
\end{definition}
Using the half-norm and the empirical distribution on a dataset $\mathcal{D}$, \citet{cohen2020learning} proves an upper (and lower) bound which is optimal with high probability. 
\begin{theorem}
    [\citet{cohen2020learning}]\label{thm:empconcentration}
    For all $M \in \mathbb{N}, \delta \in (0,1)$ and $\mu \in \Delta_\mathbb{N}$,
    \[
        \Dtv{\Tilde{P}}{P} \leq \frac{1}{\sqrt{M}}\|\Tilde{P}\|_{1/2}^{1/2} + 3\sqrt{\frac{\log(2/\delta)}{2M}}
    \]
    holds with probability at least $1-\delta$. Furthermore, 
    \(
        \E[\Dtv{\Tilde{P}}{P}] \leq \E\left[\frac{1}{\sqrt{M}}\|\Tilde{P}\|_{1/2}^{1/2}\right].
    \)
\end{theorem}
Using this result, we can easily provide a computable, high probability upper bound on the total variation distance between the true distribution and any learned PC. 
\begin{theorem}\label{thm:empiricaldistTriangle}
    Let $P$ be the true distribution, $\Tilde{P}$ be the empirical distribution, and $\hat{P}$ be our learned PC. 
    Then, with probability $1-\delta$,
    \[
        \Dtv{P}{\hat{P}} \leq \frac{1}{\sqrt{M}}\|\Tilde{P}\|_{1/2}^{1/2} + 3\sqrt{\frac{\log(2/\delta)}{2M}} + \Dtv{\Tilde{P}}{\hat{P}}.
    \]
    This upper bound can be computed in $O(\abs{\mathcal{D}}\vert{\hat{P}}\vert)$ time, where $\abs{\mathcal{D}},\vert{\hat{P}}\vert$ are the sizes of the dataset and the PC respectively.
\end{theorem}
\begin{proof}
    The result follows from a simple application of the triangle inequality and Theorem~\ref{thm:empconcentration}. 
    \begin{align*}
        \Dtv{P}{\hat{P}} &\leq \Dtv{P}{\Tilde{P}} + \Dtv{\Tilde{P}}{\hat{P}}\\
        &\leq \frac{1}{\sqrt{M}}\|\Tilde{P}\|_{1/2}^{1/2} + 3\sqrt{\frac{\log(2/\delta)}{2M}} + \Dtv{\Tilde{P}}{\hat{P}}.
    \end{align*}
    The time complexity follows from the fact that the empirical distribution can be constructed in time $O(\abs{D})$ and the TVD between the empirical distribution and the PC can be computed in time $O( \abs{\mathcal{D}} \vert\hat{P} \vert)$ using the simple computation between $\Tilde{P}$ and $\hat{P}$ below,
    \begin{align*}
    &\Dtv{\Tilde{P}}{P} = \sum_{\x\in\mathbf{X}^{\otimes M}}\abs{\Tilde{P}(\x) - \hat{P}(\x)} + \sum_{\x \notin \X^{\otimes M}} P(\x)\\
    &= \sum_{\x \in \X^{\otimes M}}\abs{\Tilde{P}(\x)-\hat{P}(\x)}+1-\sum_{\x \in \X^{\otimes M}} P(\x).
    \end{align*}
\end{proof}

We empirically test our upper bound on the density estimation benchmark datasets~\citep{lowd2010learning, bekker2015tractable, van2012markov, larochelle2011neural} with circuits learned via the HCLT structure in PyJuice~\citep{liu2024scaling}. We compute bounds on both the train and test (see Tables~\ref{tab:train} and~\ref{tab:test} in the Appendix) sets providing the error associated with each term. While our method may return trivial bounds, we observe that on several benchmark datasets of modest size we return non-trivial bounds. All learned PCs, datasets, and code is publicly available \footnote{\url{https://github.com/jsl339/StatisticalGuaranteesPCLearning}}.

\section{Conclusion}
We have proved a lower bound on the sample complexity of learning PCs: for example, any learning algorithm requires $\Omega(\frac{S + \log(1/\delta)}{\epsilon})$ samples to ensure a KL divergence within $\epsilon$ with probability at least $1-\delta$, as well as similar results for Hellinger and total variation distance. This lower bound holds even for deterministic PCs with known structure, whose maximum-likelihood parameters can be estimated in closed form. Furthermore, we showed that the parameters of deterministic and decomposable PCs are efficiently PAC learnable with sample complexity upper bounds matching the aforementioned lower bounds. Using a tournament based approach, we are able to use this technique for non-deterministic PCs in the case of the Hellinger distance and total variation distance, proving an upper bound of $O(\frac{S\log(S) + S\log(1/\delta)}{\epsilon^4} + \frac{\log(1/\delta)}{\epsilon^2})$ for both distances. This proves that PCs of any graph are PAC learnable for discrete random variables. 

There are still a number of future directions to explore. 
Notably, there remains a gap between the lower and upper bounds on the sample complexity of decomposable PC parameter learning, as well as efficient PAC learnability.
Sample complexity upper and lower bounds for PCs of unknown structures is yet unknown to the best of our knowledge. Lastly,  tight sample complexity bounds for PCs over continuous variables in the general case would be highly impactful, with a possible route via extending the results for tree-structured sum-product networks over Gaussian leaves~\citep{aden-ali_sample_2020} to DAG-structured PCs.

\bibliography{aaai2027}

\clearpage
\setcounter{secnumdepth}{2}
\appendix

\section{Derivation of Lemma~\ref{lem:Assouad} from Assouad's Lemma}\label{apx:assouad}

Here we provide a detailed derivation of how Lemma~\ref{lem:Assouad} can be derived from the statement of Assouad's lemma seen in \citep{tsybakovNonparameteric}. Furthermore, we show that Lemma~\ref{lem:Assouad} differs from that of \citep{canonne2020survey} by a factor of $1/2$ as the total variation distance satisfies the triangle inequality, while the squared Hellinger only satisfies $\Dh{P}{Q}^2 \leq 2(\Dh{P}{Z}^2 + \Dh{Z}{Q}^2).$
\begin{lemma}
    [Assouad's Lemma \citep{tsybakovNonparameteric}]
    Let $\br{0,1}^r$ be the set of all binary sequences of length $r$. Let $\br{P_v}_{v \in \br{0,1}^r} \subseteq \C$ be a set of $2^r$ probability measures, and let the corresponding expectations be denoted by $\E_v$. Then, 
    \begin{align*}
        &\inf_{\hat{v}} \max_{v \in \br{0,1}^r} \E_v[H(v,\hat{v})] \\
        &\geq \frac{r}{2} \min_{v,v': H(v,v') = 1} \inf_\psi \bigg(P_v(\psi \neq 0) + P_{v'}(\psi \neq 1)\bigg)
    \end{align*}
    where $H(v,v')$ is the Hamming distance between $v$ and $v'$, $\inf_{\hat{v}}$ denotes the infimum over all estimators taking values in $\br{0,1}^r$ and where $\inf_\psi$ denotes the infimum over all tests taking values in $\br{0,1}$.
\end{lemma}

\citet{tsybakovNonparameteric} makes the following observation in the continuous case:
\(\inf_\psi \left(P_v(\psi \neq 0) + P_{v'}(\psi \neq 1)\right) = \int \min(dP_v,dP_{v'}).\) 
Thus for the discrete case, this is equivalent to: $\inf_\psi \left(P_v(\psi \neq 0) + P_{v'}(\psi \neq 1)\right) = \sum_\x \min(P_v(\x),P_{v'}(\x))$. Using Le Cam's inequalities, we have:
\[\sum_\x \min(P_v(\x),P_{v'}(\x)) \geq \frac{1}{2}(\sum_\x \sqrt{P_v(\x)P_{v'}(\x)})^2.\] 

Note that Lemma~\ref{lem:Assouad} uses $v,v' \in \br{-1,1}^r$ instead of $v, v' \in \br{0,1}^r$ and adds the following conditions:
\begin{enumerate}[noitemsep]
    \item[(1)] For all $v,v' \in \br{-1,1}^r$, the distance between $P_v,P_{v'}$ is at least proportional to the Hamming distance: $\Dh{P_v}{P_{v'}}^2 \geq \alpha H(v,v')$.
    \item[(2)] For all $v,v' \in \br{-1,1}^r$ with $ H(v,v') = 1$, the squared Hellinger distance of $P_v,P_{v'}$ is small: $\Dh{P_v}{P_{v'}}^2 \leq \beta$.
\end{enumerate}
The squared Hellinger distance can be expressed as $\Dh{P_{v}}{P_{v'}}^2 = 1 - \sum_{\x}\sqrt{P_v(\x)P_{v'}(\x)}$. 
As such, if condition 2 is satisfied then we have that $\sum_{\x}\sqrt{P_v(\x)P_{v'}(\x)} > (1-\beta)$. If we take this over $M \geq 1$ samples, we then have $\sum_{\x}\sqrt{P^{\otimes M}_v(\x)P^{\otimes M}_{v'}(\x)} \geq (1-\beta)^M$. Next, using Condition 2,
\begin{align*}
        &\inf_{\hat{P}\in A} \max_{P \in \br{P_v}_{v \in \br{-1,1}^r}} \E_v[H(v,\hat{v})]\\
        &\geq \frac{r}{4} \bigg(\sum_{\x}\sqrt{P^{\otimes M}_v(\x)P^{\otimes M}_{v'}(\x)}\bigg)^2
        \geq \frac{r}{4} (1-\beta)^{2M}.
\end{align*}
Finally, by Condition 1, and the fact that the squared Hellinger satisfies the weak triangle inequality $\Dh{P_v}{P_{v'}}^2 \leq 2(\Dh{P_v}{\hat{P}}^2 + \Dh{\hat{P}}{P_{v'}}^2)$, we can say:
\begin{align*}
        &\inf_{\hat{P} \in A} \max_{P \in \br{P_v}_{v \in \br{-1,1}^r}} \E_v[\Dh{\hat{P}}{P}^2]\\
        &\geq \frac{\alpha r}{8} (1-\beta)^{2M}.
\end{align*}
Since $\br{P_v}_{v \in \br{-1,1}^r} \subseteq \C$, this implies exactly the following statement of Lemma~\ref{lem:Assouad}: 
\begin{align*}
        \inf_{\hat{P}} \max_{P \in \C} \E_{P}[\Dh{\hat{P}}{P}^2] \geq \frac{\alpha r}{8} (1-\beta)^{2M}.
\end{align*}

\section{Proofs}
\subsection{Proof of Proposition~\ref{prop:construction}}\label{apx:construction}

Let $\mathcal{G}$ be a deterministic PC structure of size $S$. 
W.l.o.g.\ suppose every sum node has an even number of children, and assign an index in $[1,S/2]$ to every pair of edges of the same sum node.
We parameterize the PC to define $P_v$ for each $v=\{-1,1\}^{S/2}$, where for every sum node $n$, the parameters for half of its edges will be perturbed as $\theta_{n,c}[v] = \frac{1}{\abs{\ch(n)}} + v_{n,c}\tau_n$, and the parameters for their counterparts $\theta_{n,c}[v] = \frac{1}{\abs{\ch(n)}} - v_{n,c}\tau_n$. We set $\tau_n = \sqrt{\frac{\eta}{\abs{\ch(n)}P(\gamma_n)} - \frac{\eta^2}{4P(\gamma_n)^2}}$ and $\eta \leq \min \frac{P(\gamma_n)}{16\abs{\ch(n)}D^2}$ where $D$ is the maximum number of sum nodes in a tree shaped path. Recall that for a deterministic PC, evaluating on an instantiation $\x$ is equivalent to taking the product of the parameters in the tree defined by $\x$, $P_v(\x) = \prod_{(n,c) \in \tree(\x)} \theta_{n,c}[v]$ and here we use $P(\gamma_n)$ as the probability of reaching node $n$ under the uniform circuit.

Note that for every edge $(n,c)$: if $v_{n,c}= v'_{n,c}$ then $\theta_{n,c}[v]\cdot \theta_{n,c}[v'] = (\frac{1}{\abs{\ch(n)}} \pm v_{n,c}\tau_n)^2$ where the sign $\pm$ is determined by whether it is a perturbed or dependent edge; if $v_{n,c}\neq v'_{n,c}$ then $\theta_{n,c}[v]\cdot \theta_{n,c}[v'] = (\frac{1}{\abs{\ch(n)}} + \tau_n) (\frac{1}{\abs{\ch(n)}} - \tau_n) = \frac{1}{\abs{\ch(n)}^2} - \tau_n^2$. Then we have:
\begin{align}
& \Dh{P_v}{P_{v'}}^2 = 1 - \sum_{\x} \sqrt{P_v(\x)P_{v'}(\x)} \nonumber\\
& = 1 - \sum_{\x} \sqrt{\prod_{n,c} (\theta_{n,c}[v]\theta_{n,c}[v'])^{\I((n,c) \in \tree(\x))}} \nonumber\\    
&= 1 - \sum_{\x} \Bigg[ \prod_{n,c: v_{n,c}=v'_{n,c}}(\frac{1}{\abs{\ch(n)}} \pm v_{n,c}\tau_n)^{\I((n,c) \in \tree(\x))}\nonumber\\
& \qquad \cdot \prod_{n,c: v_{n,c} \neq v'_{n,c}} \sqrt{\bigg(\frac{1}{\abs{\ch(n)}^2} - \tau_n^2\bigg)^{\I((n,c) \in \tree(\x))}} \Bigg] \label{eq:dh-bound}
\end{align}
For given $v,v'$, we can parameterize the same PC structure using the following: 
\begin{align*}
    \psi_{n,c} = \begin{cases}
        \frac{1}{\ch(n)} + v_{n,c}\tau_n & \textit{ if } v_{n,c} = v_{n,c}',\\ \\
         \sqrt{\frac{1}{\ch(n)^2} - \tau_n^2} & \textit{ if } v_{n,c} \neq v_{n,c}'.
    \end{cases}
\end{align*}
Next, we will normalize $\psi_{n,c}$ to define a valid probability distribution. To that end, we first compute the normalization constant for each $n$ as follows:
\begin{align}
    &\sum_{c \in \ch(n)} \psi_{n,c} = \frac{1}{2} \sum_{c : v_{n,c} = v_{n,c}'} (\frac{1}{\abs{\ch(n)}} + \tau + \frac{1}{\abs{\ch(n)}} - \tau) \nonumber\\
    & \qquad + \sum_{c : v_{n,c} \neq v_{n,c}'} \sqrt{\frac{1}{\ch(n)^2} - \tau_n^2} \nonumber\\
    &= \sum_{c : v_{n,c} = v_{n,c}'} \frac{1}{\abs{\ch(n)}} + \sum_{c : v_{n,c} \neq v_{n,c}'} \sqrt{\frac{1}{\ch(n)^2} - \tau_n^2} \nonumber\\
    &= \sum_{c : v_{n,c} = v_{n,c}'} \frac{1}{\abs{\ch(n)}} + \sum_{c : v_{n,c} \neq v_{n,c}'} \left( \frac{1}{\abs{\ch(n)}} - \rho_n \right) \nonumber\\
    &= 1 - \sum_{c : v_{n,c} \neq v_{n,c}'} \rho_n \nonumber\\
    &= 1 - \rho_nH_n(v,v'), \label{eq:rho}
\end{align}
where $\rho_n = \frac{2}{\abs{\ch(n)}} - 2\sqrt{\frac{1}{\ch(n)^2} - \tau_n^2} =  \frac{\eta}{P(\gamma_n)}$ to derive Equation~\ref{eq:rho}, and $H_n$ refers to the Hamming distance over only the indices corresponding to edges of $n$.
Then, for each sum node $n$, we define a normalized distribution $ \Tilde{P}_{n,c}$ as follows: 
\begin{align*}
    \Tilde{\theta}_{n,c} = \begin{cases}
        \frac{\frac{1}{\abs{\ch(n)}} + v_n\tau_n}{1 - \rho_nH_n(v,v')} & \textit{ if } v_{n,c} = v_{n,c}', \\ \\
         \frac{\sqrt{\frac{1}{\ch(n)^2} - \tau_n^2}}{1 - \rho_nH_n(v,v')} & \textit{ if } v_{n,c} \neq v_{n,c}'.
    \end{cases}
\end{align*}
Therefore, the squared Hellinger distance via Equation~\ref{eq:dh-bound} can be expressed as:
\begin{align}
    &1 - \sum_{\x}  \prod_{(n,c) \in \tree(\x)} \psi_{n,c} \nonumber\\
    &= 1 - \sum_{\x}  \prod_{(n,c) \in \tree(\x)} \tiltheta_{n,c}(1-\rho_n H_n)\nonumber\\
    &= \E_{t \sim \TP}[1 - \prod_{n \in t} (1-\rho_nH_n)]. \label{eq:dh-bound-ptilde}
\end{align}
We lower- and upper-bound the above using the fact that we set $\rho_n H_n \in [0,1]$ and Weierstrass' inequality, respectfully, as our choice of $\tau_n$ ensures the necessary conditions to apply them---$W  = \max_t \sum_{n \in t} \rho_nH_n \leq \frac{D}{32D^2} = \frac{1}{64D}$ and $\rho_n H_n \leq 1$:
\begin{equation*}
    \frac{1-e^{-W}}{W}\sum_{n \in t}\rho_nH_n \leq 1 - \prod_{n \in t} (1-\rho_nH_n)
    \leq \sum_{n \in t} \rho_n H_n.
\end{equation*}

As $W < 1$, we can improve this to the following:
\begin{equation*}
    (1-\frac{1}{64D})\sum_{n \in t}\rho_nH_n \leq 1 - \prod_{n \in t} (1-\rho_nH_n)
    \leq \sum_{n \in t} \rho_n H_n.
\end{equation*}

Thus, we can bound the expectation in Equation~\ref{eq:dh-bound-ptilde} from below using $\E_{t\sim \TP}[(1-\frac{1}{64D})\sum_{n \in t}\rho_nH_n] \geq \frac{63}{64}\sum_{n}\rho_nH_n \E_{t\sim \TP}[\I(n \in t)] = \frac{63}{64}\sum_{n}\rho_nH_n \TP(\gamma_n)$
and similarly from above using $\E_{t\sim \TP}[\sum_{n \in t} \rho_n H_n] = \sum_{n} \rho_n H_n \TP(\gamma_n)$.

Next, remember that $\tiltheta_{n,c} = \frac{\psi_{n,c}}{\sum_{c'}\psi_{n,c'}}$ and 
\[
    \frac{1}{\abs{\ch(n)}} - \tau_n \leq \psi_{n,c} \leq \frac{1}{\abs{\ch(n)}} + \tau_n.
\]
Since, $H_n \leq \frac{\abs{\ch(n)}}{2}$, we have that
\[
    \sum_{c'}\psi_{n,c'} = 1 - \rho_nH_n \geq 1 - \frac{\abs{\ch(n)}}{2}\rho_n = \sqrt{1-\abs{\ch(n)}^2\tau_n^2}
\]

This allows us to say that:
\[
    \frac{1}{\abs{\ch(n)}} - \tau_n \leq \tiltheta_{n,c} \leq \frac{1}{\abs{\ch(n)}(1-\abs{\ch(n)}\tau_n)}.
\]
Now, by our choice of $\eta$, we know that
$\rho_n \leq \frac{1}{16\abs{\ch(n)}D^2}$, which implies $\tau_n \leq \sqrt{\frac{1}{\abs{\ch(n)}}\rho_n} \leq \frac{1}{4\abs{\ch(n)}D}$ and $\tau_n\abs{\ch(n)} \leq \frac{1}{4D}$. Therefore,
\begin{align*}
     1 - \frac{1}{4D} &\leq \frac{\tiltheta_{n,c}}{1/\abs{\ch(n)}} \leq \left(1 - \frac{1}{4D}\right)^{-1}
\end{align*}
Since $D$ is the maximum number of edges in a tree-shaped path through the circuit, we have that any root to node path contains at most $D$ factors, which yields, 
\[
    \left(1 - \frac{1}{4D}\right)^D P(\gamma_n) \leq \TP(\gamma_n) \leq \left(1 - \frac{1}{4D}\right)^{-D}P(\gamma_n)
\]
Taking Bernoulli's inequality gives us $\left(1 - \frac{1}{4D}\right)^D \leq 1 - \frac{D}{4D} = \frac{3}{4}$ and thus
\[
    \frac{3}{4}P(\gamma_n) \leq \TP(\gamma_n) \leq \frac{4}{3}P(\gamma_n).
\]

Furthermore, $w = \max_t \sum_{n \in t} \rho_nH_n \leq \frac{D}{32D^2} = \frac{1}{32D}$ and since $1 - \frac{w}{2} \geq 1 - \frac{1}{64D} \geq \frac{63}{64}$ we can put this together to prove the constructed family meets the conditions of Lemma~\ref{lem:Assouad}.

\begin{align*}
    \Dh{P_v}{P_{v'}}^2 &\geq (1 - \frac{w}{2}) \sum_{n}\rho_nH_n P_v(\gamma_n))\\
    &\geq \frac{63}{64}\frac{3}{4}\sum_{n,c}\frac{\eta}{P(\gamma_n)}H_n P_v(\gamma_n)) \\
    &\geq \frac{189}{256}\eta H(v,v')
\end{align*}
and

\begin{align*}
    \Dh{P_v}{P_{v'}} &\leq \sum_{n} \rho_n \TP(\gamma_n) 
    \leq \frac{\eta}{P(\gamma)}\TP(\gamma_n)) 
    \leq \frac{4}{3} \eta
\end{align*}
\qed

\subsection{Proof of Lemma \ref{lemma:reverseKLPC}}\label{apx:proofRKLD}
Suppose that $\hat{P}$ and $P$ are deterministic and decomposable PCs with the same structure over variables $\Y$. Then for all $y$:
    \begin{align*}
    \log(\frac{\hat{P}(\y)}{P(\y)}) &= \log(\frac{\prod_{e \in \tree(\y)} \hat{\theta}_e}{\prod_{e\in\tree(\y)} \theta_e})
    = \sum_{e \in \tree(\y)} \log(\frac{\hat{\theta}_e}{\theta_e})\\
    &= \sum_{n,c}\log(\frac{\hat{\theta}_{n,c}}{\theta_{n,c}}) \I((n,c) \in \tree(\y))
    \end{align*}
Next, use the definition of the reverse KL divergence:
\begin{align*}
    &\Dkl{\hat{P}}{P} = \sum_{\y} \hat{P}(\y)\log(\frac{\hat{P}(\y)}{P(\y)})\\
    &= \sum_{\y} \hat{P}(\y) \sum_{n,c} \log(\frac{\hat{\theta}_{n,c}}{\theta_{n,c}}) \I((n,c) \in \tree(\y))\\
    &= \sum_{n,c} \log(\frac{\hat{\theta}_{n,c}}{\theta_{n,c}}) \sum_{\y} \hat{P}(\y)\I((n,c) \in \tree(\y))\\
    &= \sum_{n,c} \log(\frac{\hat{\theta}_{n,c}}{\theta_{n,c}}) \hat{P}(\I((n,c)\textit{ is active })\\
\end{align*}
Our event in this case can be seen as two parts: (1) the probability of the context of the node, and (2) the probability of the context of the child. Thus, 
\begin{align*}
    \sum_{n,c} \log(\frac{\hat{\theta}_{n,c}}{\theta_{n,c}}) \hat{P}(\gamma_n \wedge \gamma_c)
    &= \sum_{n,c} \log(\frac{\hat{\theta}_{n,c}}{\theta_{n,c}}) \hat{P}(\gamma_n) \hat{P}(\gamma_c \vert \gamma_n)\\
    &= \sum_{n,c} \log(\frac{\hat{\theta}_{n,c}}{\theta_{n,c}}) \hat{P}(\gamma_n)\hat{\theta}_{n,c}\\
    &= \sum_n \hat{P}(\gamma_n) \Dkl{\hat{P}_c}{P_c}.
\end{align*}
\qed

\section{Experimental Results}
We compute our proposed bound for $\delta = 0.1$ on the train and test sets of density estimation benchmark datasets. Tables~\ref{tab:train} and~\ref{tab:test} show that we are able to achieve a number of non-trivial results, and the bound improving as the number of samples grows larger. 
\begin{table}[htbp]
\centering
\scriptsize
\setlength{\tabcolsep}{3pt}
\begin{tabular}{lrrrrrr}
\hline
\shortstack{Dataset} & Vars & $M$ & $S$ & $\Dtv{P}{\Tilde{P}}$ & $\Dtv{\Tilde{P}}{\hat{P}}$ & $\Dtv{P}{\hat{P}}$ \\
\hline
accidents & 111 & 12758 & 1612 & 1.032414 & 0.999995 & 2.032409 \\
ad & 1556 & 2461 & 21332 & 0.808336 & 0.882735 & 1.691071 \\
baudio & 100 & 15000 & 1748 & 1.028706 & 0.999650 & 2.028356 \\
bbc & 1058 & 1670 & 15332 & 1.060031 & 1.000000 & 2.060031 \\
bnetflix & 100 & 15000 & 1652 & 1.029894 & 1.000000 & 2.029894 \\
book & 500 & 8700 & 7172 & 1.021469 & 0.988659 & 2.010128 \\
c20ng & 910 & 11293 & 12052 & 1.031496 & 1.000000 & 2.031496 \\
connect4 & 126 & 16000 & 1780 & 1.029027 & 0.992987 & 2.022014 \\
cr52 & 889 & 6532 & 11612 & 1.027842 & 0.999978 & 2.027820 \\
cwebkb & 839 & 2803 & 11900 & 1.056962 & 1.000000 & 2.056962 \\
dna & 180 & 1600 & 3268 & 1.071262 & 1.000000 & 2.071262 \\
jester & 100 & 9000 & 1636 & 1.034563 & 0.999952 & 2.034515 \\
kdd & 64 & 180092 & 1108 & 0.065394 & 0.104025 & \textbf{0.169419} \\
kosarek & 190 & 33375 & 2580 & 0.425928 & 0.369643 & \textbf{0.795570} \\
moviereview & 1001 & 1600 & 15020 & 1.090692 & 1.000000 & 2.090692 \\
msnbc & 17 & 291326 & 236 & 0.090154 & 0.340814 & \textbf{0.430968} \\
mushrooms & 112 & 2000 & 1652 & 1.082100 & 0.996632 & 2.078732 \\
nips & 500 & 400 & 8212 & 1.183581 & 1.000000 & 2.183581 \\
nltcs & 16 & 16181 & 260 & 0.304177 & 0.237589 & \textbf{0.541767} \\
ocr\_letters & 128 & 32152 & 2820 & 0.987693 & 0.999993 & 1.987686 \\
plants & 69 & 17412 & 1324 & 0.593797 & 0.854377 & 1.448173 \\
pumsb\_star & 163 & 12262 & 2396 & 1.021908 & 0.999073 & 2.020980 \\
rcv1 & 150 & 40000 & 1972 & 0.959677 & 0.999621 & 1.959298 \\
tmovie & 500 & 4524 & 7876 & 0.960019 & 0.944952 & 1.904971 \\
tretail & 135 & 22041 & 1948 & 0.573789 & 0.492148 & 1.065937 \\
voting & 1359 & 1214 & 16572 & 0.854437 & 1.000000 & 1.854437 \\
\hline
\end{tabular}
\caption{Upper bounds for the train split with $\delta = 0.1$.}
\label{tab:train}
\end{table}

\begin{table}[htbp]
\centering
\scriptsize
\setlength{\tabcolsep}{3pt}
\begin{tabular}{lrrrrrr}
\hline
\shortstack{Dataset} & Vars & $M$ & $S$ & $\Dtv{P}{\Tilde{P}}$ & $\Dtv{\Tilde{P}}{\hat{P}}$ & $\Dtv{P}{\hat{P}}$ \\
\hline
accidents & 111 & 2551 & 1612 & 1.072695 & 0.999999 & 2.072694 \\
ad & 1556 & 491 & 21332 & 1.003393 & 0.888762 & 1.892155 \\
baudio & 100 & 3000 & 1748 & 1.066839 & 0.999937 & 2.066776 \\
bbc & 1058 & 330 & 15332 & 1.196791 & 1.000000 & 2.196791 \\
bnetflix & 100 & 3000 & 1652 & 1.067034 & 1.000000 & 2.067034 \\
book & 500 & 1739 & 7172 & 1.082414 & 0.996639 & 2.079053 \\
c20ng & 910 & 3764 & 12052 & 1.056903 & 1.000000 & 2.056903 \\
connect4 & 126 & 47557 & 1780 & 1.016836 & 0.979539 & 1.996376 \\
cr52 & 889 & 1540 & 11612 & 1.080821 & 0.999996 & 2.080817 \\
cwebkb & 839 & 838 & 11900 & 1.118674 & 1.000000 & 2.118674 \\
dna & 180 & 1186 & 3268 & 1.091390 & 1.000000 & 2.091390 \\
jester & 100 & 4116 & 1636 & 1.053856 & 0.999962 & 2.053818 \\
kdd & 64 & 34955 & 1108 & 0.098416 & 0.097484 & \textbf{0.195900} \\
kosarek & 190 & 6675 & 2580 & 0.536955 & 0.431198 & \textbf{0.968153} \\
moviereview & 1001 & 250 & 15020 & 1.232214 & 1.000000 & 2.232214 \\
msnbc & 17 & 58265 & 236 & 0.169269 & 0.355389 & \textbf{0.524658} \\
mushrooms & 112 & 5624 & 1652 & 1.048959 & 0.990377 & 2.039337 \\
nips & 500 & 1240 & 8212 & 1.103795 & 1.000000 & 2.103795 \\
nltcs & 16 & 3236 & 260 & 0.499338 & 0.331435 & \textbf{0.830773} \\
ocr\_letters & 128 & 10000 & 2820 & 1.013833 & 0.999996 & 2.013829 \\
plants & 69 & 3482 & 1324 & 0.755341 & 0.881216 & 1.636557 \\
pumsb\_star & 163 & 2452 & 2396 & 1.072236 & 0.999704 & 2.071941 \\
rcv1 & 150 & 150000 & 1972 & 0.912592 & 0.999368 & 1.911960 \\
tmovie & 500 & 591 & 7876 & 1.092290 & 0.960803 & 2.053093 \\
tretail & 135 & 4408 & 1948 & 0.710520 & 0.568160 & 1.278680 \\
voting & 1359 & 350 & 16572 & 0.968925 & 1.000000 & 1.968925 \\
\hline
\end{tabular}
\caption{Upper bounds for the test split with $\delta = 0.1$.}
\label{tab:test}
\end{table}

\end{document}